\documentclass[11pt]{article}

\usepackage[preprint]{acl}
\usepackage[utf8]{inputenc}        % Encodage UTF-8
\usepackage{longtable}
\usepackage[T1]{fontenc}           % Encodage des fonts
\usepackage{amsmath}               % Mathématiques avancées
\usepackage{amssymb}               % Symboles mathématiques
\usepackage{amsfonts}              % Fonts mathématiques (pour \mathbb{R})
\usepackage{graphicx}              % Inclusion d'images
\usepackage{float}                 % Pour [H] et [t] dans figures/tables
\usepackage{booktabs}              % Beaux tableaux (pour \hline)
\usepackage{array}                 % Colonnes de tableaux avancées (pour p{})
\usepackage{multirow}              % Cellules multi-lignes (optionnel)
\usepackage{subcaption}            % Sous-figures (optionnel)
\usepackage{tabularx}
\usepackage{booktabs}
\usepackage{array}
\usepackage{hyperref}              % Liens hypertextes et références
\usepackage{url}                   % URLs formatées

\setcitestyle{authoryear,round}
\usepackage{dblfloatfix}
\usepackage{xcolor}                % Couleurs (pour highlighting)
\usepackage{algorithm}             % Algorithmes
\usepackage{algorithmic}           % Pseudo-code
\usepackage{listings}              % Code source
\usepackage{caption}               % Captions personnalisée
\usepackage{times}
\usepackage{latexsym}
\usepackage{latexsym}
\usepackage{amsmath}
\usepackage[T1]{fontenc}
\usepackage[utf8]{inputenc}

\usepackage{microtype}

\usepackage{inconsolata}

\usepackage{graphicx}

\usepackage{placeins}

\usepackage{soul}

\title{Hybrid Adversarial Defence for Natural Language Understanding Tasks}

\author{
    \textbf{Manar Abouzaid\textsuperscript{1}},
    \textbf{Yang Wang\textsuperscript{2}},
    \textbf{Chenghua Lin\textsuperscript{2}},\\
    \textbf{Stuart E. Middleton\textsuperscript{1}} \\
    \textsuperscript{1}School of Electronics and Computer Science, University of Southampton, UK \\
    \textsuperscript{2}Department of Computer Science, University of Manchester, UK \\
}

\begin{document}
\maketitle
\begin{abstract}
Large Language Models (LLMs) are vulnerable both to hallucination and adversarial manipulation. Although these problems are closely related, existing defences typically address them separately. We investigate a hybrid defence framework that combines entropy-based models, designed to reduce hallucinations, with uncertainty-based models and geometric-based models, designed to reduce vulnerability. Under in-domain tests on Natural Language Understanding datasets (FEVER, HotpotQA, CSQA, SIQA) we find our hybrid model improves both clean-task performance (up to 43.34\% increase in accuracy) and adversarial robustness (up to 64.92\% improvement in accuracy and 62.27\% reduction in attack success rate). For out-of-distribution datasets (AeroEngQA, CPIQA) we see similar adversarial robustness from our hybrid model (up to 57.14\% improvement in accuracy). For prompt injection (SafeGuard) and jailbreak detection (AdvBench, DAN) datasets our hybrid model is also very strong (up to 51\% reduction in attack success rate compared to state of the art baseline models). Overall, our results show that combining entropy, uncertainty and geometric features provides a more effective defence strategy than using any single feature alone for both in-domain and out-of-distribution tasks.
\end{abstract}

% white space before and after equations
% avoid empty line before eq otherwise it becomes a new paragraph
\setlength{\abovedisplayskip}{6pt}
\setlength{\belowdisplayskip}{6pt}

\section{Introduction}

Large Language Models (LLMs) have become an important component of modern artificial intelligence systems \citep{brown2020language, touvron2023llama}. Despite their impressive capabilities, two well-known issues continue to affect large language models: hallucination and adversarial vulnerability.

In this paper, we explore whether methods designed to address hallucination and adversarial attacks can be combined within a single framework to improve the robustness of LLMs for Natural Language Understanding (NLU) tasks. Our motivation is based on the observation that there are now several very different classes of hallucination reduction and adversarial defence model being explored in the literature, and these models very likely each have a distinct error distribution. If we were able to learn when to use each class of model we could develop a powerful hybrid adversarial defence approach. The first class of model we use is an entropy-based model motivated by \citet{yao2023llm} to reduce hallucinations. The second is an uncertainty trained model able to say `I don't know' motivated by \citet{zhang2024rtuning}. The third is a geometric-based model motivated by \citet{wang2025adversarial} which improves hidden states by removing dominant principal components. To aggregate these defences, we introduce a selector network that learns to route inputs to our three expert models. The routing decisions are based on features capturing entropy features, uncertainty estimates, and the geometric properties of the model’s internal representations. We explore both hard expert selection and soft probabilistic aggregation of defence features. We release our models\footnote{Paper github URL once paper accepted} as open source for reproducibility purposes.

We investigate the following research questions:

\begin{itemize}

\item RQ1 Can token-level in-domain adversarial defence methods (entropy-based, uncertainty-based, geometric-based) improve the base performance of NLU tasks in addition to robustness?

\item RQ2 Can hybrid adversarial defence methods outperform single adversarial defence methods for both in-domain and out-of-distribution tasks?

\item RQ3 Do token-level defence mechanisms generalize to sequence-based attacks such as prompt injection and jailbreak attempts?

\end{itemize}

This work makes the following contributions:

\begin{itemize}

\item We describe the first hybrid adversarial defence approach to use entropy, uncertainty and geometric-based features to select adversarial defence models. Our hybrid adversarial defence consistently out-performs state of the art models on both task performance and robustness across a variety of in-domain NLU datasets (FEVER, HotpotQA, CSQA, SIQA).

\item We provide a comprehensive analysis of how our hybrid adversarial defence performs in out-of-distribution and sequence-based attack problems. Our hybrid adversarial defence out-performs any single model on all out-of-distribution (AeroEngQA, CPIQA), prompt injection (SafeGuard) datasets and jailbreak (AdvBench, DAN) datasets, except for DAN where the NeMoGuard model that was explicitly optimised for this dataset was best.

\end{itemize}

\section{Related Work}

\subsection{Hallucination Reduction}

Large language models (LLMs) hallucinations \citep{zhang2025siren, ji2023survey} are responses that are confident and well written, but which contain errors such as being factually incorrect or nonsensical. Recent large-scale studies have confirmed the continued prevalence of hallucinations \citep{massenon2025my}. Research into LLM hallucination reduction has mostly focused on the areas of improving faithfulness of responses to the original input, factual accuracy of responses and consistency of responses to reduce contradictions. 

Recent work on improving LLM faithfulness includes entropy-based approaches to detect statistical anomalies in LLM responses, such as LLMLies \citep{yao2023llm}, and uncertainty-aware training to allow models to refrain from answering questions beyond a knowledge threshold, such as R-Tuning \citep{zhang2024rtuning}. Retrieval-Augmentation Generation (RAG) \citep{lewis2020retrieval} has long been used to improve LLM factual accuracy by retrieving context from trusted sources. Other methods to improve factual accuracy include use of semantic entropy metrics \citep{zubkova2025sugar} to decide when to use pre-trained LLM knowledge and when to augment it with RAG, perplexity-based metrics \citep{varshney2023stitch} to avoid unnatural out of distribution responses and knowledge injection \cite{elaraby2023halo} using expert-verified knowledge sets to fill specific factual gaps that would otherwise lead to LLM hallucinations. Recent work to improve LLM consistency includes generate-then-refine strategies \citep{dziri2021neural} using a knowledge-graph, entailment models \citep{maynez2020faithfulness} to assess quality of abstractive summarization and self-consistency approaches \citep{manakul2023selfcheck} where responses for similar concepts are expected to be similar and contain consistent facts.

Our work uses both entropy-based and uncertainty-based approaches to reduce LLM hallucinations, but we do this in the context of aggregating them with geometric-based adversarial defence which is something that has not been explored before.

\subsection{Adversarial Defence}

Adversarial defence attempts to make LLMs more robust to adversarial attacks such as malicious input perturbation, prompt injection and LLM jailbreaking. There are several adversarial attack frameworks available to test LLMs, such as TextAttack \citet{morris2020textattack}, TextFooler \citet{jin2020textfooler}, TextBugger \citet{li2018textbugger} and PWWS \citet{ren2019pwws}. Jailbreak attacks \citep{wei2023jailbroken} \citep{ding2023wolf} \citep{zhang-etal-2025-damon} \citep{fu-etal-2025-jailbreak} exploit vulnerabilities in LLMs, potentially bypassing alignment constraints whilst prompt injection attacks can bypass system instructions \citep{greshake2023promptinjection}. Recent studies have shown current LLM guardrail systems can still be compromised by attacks \citep{hackett2025}. 

Adversarial defence methods include adversarial training, perturbation control and use of regularization during training to encourage model robustness. Adversarial training involves augmenting training data with adversarial perturbations, either explicitly adding new adversarial data instances using an adversarial attack framework or implicitly augmenting training data by dynamically adjusting it during the embedding phase. Examples of implicit perturbation include \citet{gao2023-dsrm}, \citet{latorre2023} and FreeLB \citep{zhu2020freelb}. It should be noted that adversarial training is often computationally expensive as it increases training data volumes and requires more training time when fine-tuning. Perturbation control is where models aim to recognize and correct perturbations during training. Examples include the use of spell-checking \citet{alshemali2019}, synonym substitution \citet{dong2021robustnessnaturallanguageword} when perturbation sets are known a-priori and data cleaning \citet{bao-etal-2021-defending} to limit adversarial data input space. Regularization type approaches focus on rewarding models during training to encourage robustness, such as \citet{wang2021infobert}, which uses two regularizers to improve out-of-domain robustness, and \citet{bianchi2024safetytuned}, which integrates adversarial training and reinforcement learning to enhance robustness. Other approaches include PURE \citep{wang2025adversarial}, which improves robustness by removing dominant principal components in the hidden representation to create a more isotropic space, SafeLoRA \citet{hsu2024safe}, which projects LoRA weights onto a safety-focussed subspace to mitigate safety risks, and SPLoRA \citet{ao-etal-2025-safe} which prunes LoRA layers that significantly deviate from pre-trained LLM states and could introduce security vulnerabilities.

To our knowledge, no prior adversarial defence work has explored aggregating entropy-based, uncertainty-based and geometric-based methods as we do in this paper to bridge the gap between hallucination and adversarially focused approaches.

\section{Hybrid Adversarial Defence Framework}

Our hybrid adversarial defence framework, shown in Figure~\ref{fig:hybrid_adv_defence}, consists of three individual adversarial defence models and a choice of two aggregation methods. The input queries are first sent to individual defence models, an entropy-based, uncertainty-based and geometry-based adversarial defence model. The aggregation methods then exploit classes of features available from each individual model to train a routing algorithms, which ultimately provides an adversarial defence decision to accept or reject an input query.

% \FloatBarrier % prevent later figs appearing before this

\subsection{Entropy-based Model}

Our entropy-based adversarial defence model is motivated by the work of \citet{yao2023llm}, which uses gradient-based adversarial generation. For each token position $i$, the original method computes gradients with respect to token embeddings, identifying top-$k$ replacement tokens with complexity $O(T \times L \times K \times V)$, where $T$ is the number of epochs, $L$ is the input length, $K$ is the number of top-$k$ candidates, and $V$ is the vocabulary size.

Instead of the original gradient-based generator, we produce adversarial examples using the TextAttack framework of \citet{morris2020textattack}. This pragmatic choice greatly improves runtime speed, making large-scale evaluation feasible, and yields adversarial inputs that are effective for testing of entropy-based detection. Appendix~\ref{appendix:attack_strategies} summarizes the attack recipes employed.

We use an entropy-based thresholding mechanism to reject adversarial prompts during inference. Adversarial inputs induce uncertainty, resulting in higher-entropy distributions across the vocabulary. During inference, entropy $H$ is computed \eqref{eq:1} over softmax-normalized logits of the first predicted token where $p_i$ denotes the softmax probability assigned to token $i$ in vocabulary $V$.
\begin{equation} \label{eq:1}
H = -\sum_{i=1}^{V} p_i \log p_i
\end{equation}
If the entropy exceeds a set threshold, the prompt is marked as adversarial and rejected. This provides a simple and efficient defence against adversarial prompts, without requiring model retraining. Unlike \citet{yao2023llm}, which uses a universal entropy threshold, we employ dataset-specific thresholds calculated from the average entropy observed within each domain. This improves detection robustness across different task types and datasets.

%
%\begin{figure*}[h]
%  \centering
%  \includegraphics[width=1.0\textwidth]{latex/IMG1.png}
%  \caption{Entropy-based Model: Deterministic adversarial generation and entropy-based detection mechanism.}
%  \label{fig:llmlies}
%\end{figure*}
%

\subsection{Uncertainty-based Model}

Our uncertainty-based adversarial defence model is motivated by \citet{zhang2024rtuning}'s Refusal-Aware Instruction Tuning (R-Tuning) methodology, which addressed hallucinations by training models to recognize knowledge boundaries and express appropriate uncertainty. Given pre-trained parametric knowledge ($P$) and instruction-tuned knowledge ($I$), the model's reliable knowledge space ($P \cap I$) is where confident responses are expected, while questions outside this intersection ($I \setminus P$) are uncertain and should be refused.

The training dataset $D = \{(q_1, a_1), \ldots, (q_n, a_n)\}$ is partitioned \eqref{eq:2} into certain ($D_1$) and uncertain ($D_0$) subsets based on the pre-trained model's parametric knowledge. For each question-answer pair $(q_i, a_i)$, the pre-trained model $M_0$ generates prediction $\hat{a}_i$ without fine-tuning.
\begin{align} \label{eq:2}
\hat{a}_i &= M_0(q_i) \\
\nonumber (q_i, a_i) &\in D_1 \text{ if } \hat{a}_i = a_i \text{ (certain)} \\
\nonumber (q_i, a_i) &\in D_0 \text{ if } \hat{a}_i \neq a_i \text{ (uncertain)}
\end{align}
\begin{align} \label{eq:3}
\text{Template} &= \text{``Q: \{Question\}, A: \{Answer\}.} \nonumber \\
&\quad \text{\{Uncertainty\_Prompt\}''}
\end{align}
A refusal-aware training dataset is constructed using a standardized template \eqref{eq:3} that combines questions, answers, and uncertainty assessment where $\text{Uncertainty\_Prompt}$ = ``Are you sure you accurately answered the question based on your internal knowledge?''. Differential responses are appended based on knowledge certainty classification:
\begin{itemize}
    \item Certain Data ($D_1$): Append ``I am sure''
    \item Uncertain Data ($D_0$): Append ``I am unsure''
\end{itemize}

This padding strategy maintains ground-truth labels while explicitly adding confidence labels. The model is then fine-tuned using cross-entropy loss calculated using response and uncertainty tokens.

%
%\begin{equation}
%L = -\frac{1}{T} \sum_{i=1}^{T} \log P(t_i|t_1, t_2, \ldots, t_{i-1})
%\end{equation}
%

During inference, a two-stage process extracts both predictions and certainty measures. First, the model generates an answer. Then, an uncertainty assessment prompt is appended to evaluate the model's certainty about its response. The certainty score is computed by extracting the model's logits for ``sure'' and ``unsure'' tokens and normalizing \eqref{eq:4} where $p_{\text{sure}}$ and $p_{\text{unsure}}$ are the probabilities assigned to the respective tokens by the model's output distribution. This provides a calibrated uncertainty measure enabling explicit knowledge boundary recognition.
\begin{equation} \label{eq:4}
c = \frac{p_{\text{sure}}}{p_{\text{sure}} + p_{\text{unsure}}}
\end{equation}

%
%\begin{figure*}[h]
%    \centering
%    \includegraphics[width=1.0\textwidth]{2.png}
%    \caption{Uncertainty-based Model: Knowledge-aware uncertainty training for refusal-aware fine-tuning.}
%    \label{fig:rtuning}
%\end{figure*}
%

\subsection{Geometric-based Model}

Our geometric model draws inspiration from the \textsc{PuRe} algorithm proposed by \citet{wang2025adversarial}, which increases adversarial robustness via Principal Component Removal (PCR) at the instance level, in an effort to reduce vulnerability to adversarial attacks. The goal is to reduce the influence of certain directions in the representation space by discarding the principal components responsible for the majority of variance. For instance-level representations $X \in \mathbb{R}^{n \times d}$ where $n$ is the sequence length and $d$ is the embedding dimension, we apply Singular Value Decomposition (SVD) as $X = U \Sigma V^\top$, where $U$ and $V$ represent the left and right singular vectors, respectively, and $\Sigma$ is a diagonal matrix comprising the singular values.

The top-1 principal component $v_1$, or the largest singular vector, is discarded as described in \eqref{eq:5}.
\begin{equation} \label{eq:5}
X \leftarrow X - (Xv_1) v_1^\top
\end{equation}
By removing the rank-1 component corresponding to the largest singular value, the process eliminates dominating directions representing high-frequency token sequences and frequently occurring syntactic structures utilized by the adversaries.

In order to ensure efficient implementation, we use randomized Singular Value Decomposition (SVD) \citep{halko2011}. This technique relies on Gaussian random projections to estimate the principal components and achieves significant savings in computation time compared to traditional SVD.

Following the removal of the principal components, we apply mean pooling of token-based representations using a parameter-free form of self-attention originally proposed by \citet{zhai2023}, which serves to aggregate tokens into a sentence representation. The advantage of this method is that it efficiently captures patterns at the sentence-level in linear time with no addition of any trainable parameter. This prevents the model from overfitting and provides the model with a richer semantic representation.

Unlike entropy and uncertainty-based models, which detect adversarial inputs and selectively refuse to answer, \textsc{PuRe} enhances robustness by transforming the embedding space itself, making representations more isotropic and less susceptible to adversarial perturbations.

%
%\begin{figure*}[h]
%    \centering
%    \includegraphics[width=1.0\textwidth]{latex/IMG3.png}
%    \caption{Geometric-based Model: Instance-level Principal Component Removal to reduce adversarial susceptibility.}
%    \label{fig:PURE}
%\end{figure*}
%

\begin{figure*}[t]
    \centering    \includegraphics[width=\textwidth]{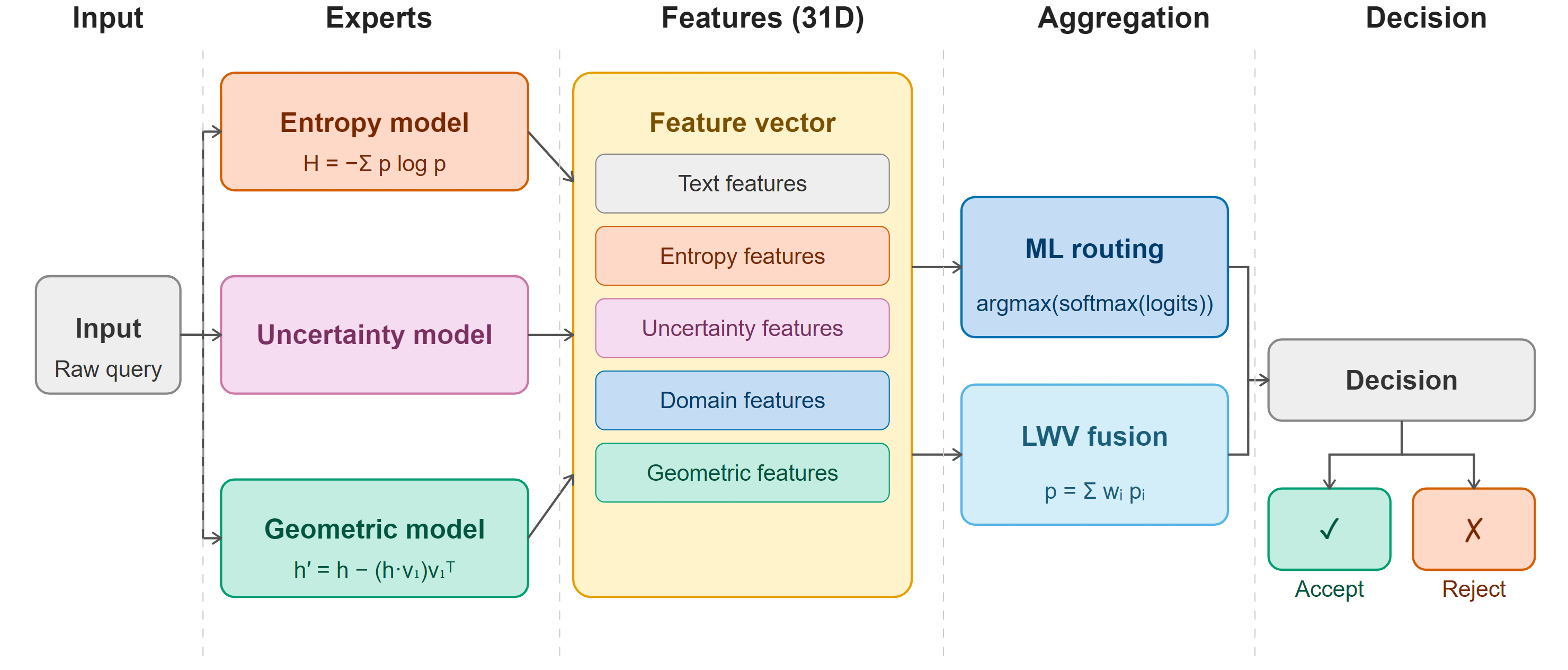}
    \caption{Hybrid Adversarial Defence Framework}
    \label{fig:hybrid_adv_defence}
\end{figure*}

%\FloatBarrier % prevent defence model fig appearing before the experiment section

\subsection{Hybrid Adversarial Defence}
\label{sec:hybrid_defence}

%
%\begin{figure*}[h]
%    \centering
%    \includegraphics[width=1.0\textwidth]{latex/hybrid2_structure.png}
%    \caption{Hybrid Adversarial defence: Multi-Expert Framework}
%    \label{fig:hybrid_system}
%\end{figure*}
%

Our hybrid adversarial defence model integrates entropy-based, uncertainty-based and geometric-based models using a trainable feature-based routing mechanism. Figure \ref{fig:hybrid_adv_defence} illustrates the overall architecture. We explore two aggregation strategies that share a common feature representation but differ in how decisions are made.

\textbf{ML Routing:} A feed-forward network $f_{\text{route}}(x)$ with a hidden layer of size 512 and ReLU activation takes the feature vector $x \in \mathbb{R}^{31}$ as input and outputs a discrete expert selection where $k \in \{rt, llm, pure\}$ denotes the selected expert. The final prediction is taken directly from the selected expert \eqref{eq:6}.
\begin{align} \label{eq:6}
k &= \arg\max \, f_{\text{route}}(x)
\end{align}

\textbf{LWV Fusion:} A second feed-forward network $f_{\text{weight}}(x)$, with the same architecture as the routing network, produces a distribution over experts where $p_{rt}$, $p_{llm}$, and $p_{pure}$ are the predicted probabilities from each expert \eqref{eq:7}.
\begin{align} \label{eq:7}
[w_{rt}, w_{llm}, w_{pure}] &= \text{softmax}(f_{\text{weight}}(x)) \\
p_{\text{final}} &= w_{rt} \cdot p_{rt} \nonumber \\
&\quad + w_{llm} \cdot p_{llm} \nonumber \\
&\quad + w_{pure} \cdot p_{pure} \nonumber
\end{align}
Both models use identical architectures to ensure a fair comparison between discrete routing (hard selection) and continuous weighting (soft aggregation). The key difference lies in the decision mechanism: hard routing selects a single expert, while soft aggregation combines all experts through learned weights.

We extract 31 features from the three expert models across six categories: 6 textual statistics, 6 entropy features, 5 uncertainty features, 3 geometric features, 5 domain encoding and 6 cross-expert patterns. Full details of the features extracted are provided in Appendix~\ref{appendix:features}, but each is encoded as a scalar within a feature vector we used as input to our aggregation model.

Entropy-based features directly detect adversarial inputs by monitoring output entropy patterns, with high entropy in first-token predictions triggering refusal. Uncertainty-based features detects knowledge boundary violations through calibrated confidence scores, allowing refusal of inputs where the model is uncertain. Geometric-based features include indicators of purification magnitude, norm ratios, and singular value dominance which can reveal how much the embedding space was modified during purification.

\section{Baseline Models}
\subsection{FreeLB - Adversarial Defence}

FreeLB by \citet{zhu2020freelb} is an adversarial training baseline that strengthens model robustness by perturbing word embeddings during fine-tuning, accumulating parameter gradients across multiple inner perturbation steps, then appling a single consolidated update. Our configuration follows the original work's hyperparameters reported in the original paper. The perturbation budget $\epsilon$ ranges from 0.15 to 0.6 depending on the dataset, with ascent step sizes $\alpha$ between 0.025 and 0.15. All experiments use $K=3$ inner steps with gradients normalized.

% \begin{figure*}[h]
%    \centering
%    \includegraphics[width=1.0\textwidth]{latex/freelb_structure.png}
%    \caption{Baseline Model - FreeLB}
%    \label{fig:freelb}
%\end{figure*}

\subsection{ProtectAI v1 - Prompt Injection Detection}
ProtectAI v1  is a DeBERTa-v3-base classifier trained on a dataset that integrates prompt injection and benign prompts\citep{protectai2023v1}. Approximately 30\% of the prompts in this dataset are injections and 70\% are benign. This classifier takes an input prompt and classifies it as either an injection or non-injection. ProtectAI v1 is specifically designed to detect prompt injections; it was not trained to detect jailbreak attacks \citep{hackett2025}.

\subsection{NeMoGuard: Jailbreak Detection Baseline}
\citet{galinkin2024} introduce NeMoGuard, a jailbreak detection framework combining pretrained text embeddings with a Random Forest classifier. The model was trained on a set of 17,085 aggregated examples from three separate datasets: the DAN dataset \citep{shen2024}, containing 1,405 in-the-wild jailbreak prompts; the GARAK dataset, containing 126 algorithmically generated jailbreaks using the AutoDAN and TAP algorithms; and the jackhhao dataset, containing 666 jailbreak prompts. NeMoGuard represents a specialized jailbreak baseline.

\subsection{Task Datasets}
\label{dataset_spec}

We evaluate all baseline models and hybrid adversarial defence method on a wide range of datasets covering different aspects of robustness in large language models (LLMs). For NLU datasets (Table~\ref{tab:exp1_datasets}), FEVER \citep{thorne2018fever} and HotpotQA \citep{yang2018hotpot} assess factuality and multi-step reasoning, while CommonsenseQA \citep{talmor2019commonsense} and Social IQA \citep{sap2019social} evaluate commonsense reasoning. For out-of-distribution datasets. AEROENG \citep{silva2025retrieval} and CIPQA \citep{mutalik2025cpiqa} are selected as they cover scientific and technical domains not present in the NLU datasets. For instruction-level adversarial robustness, we use SafeGuard \citep{erdogan2024safeguard} to evaluate prompt injection, and AdvBench \citep{zou2023} and DAN \citep{shen2024} to evaluate jailbreak attacks.

\section{In-Domain Adversarial Defence}
\label{sec:exp1}

We evaluate the following model configurations representing baseline, individual defences and hybrid defence:
\begin{itemize}
    \item \textbf{Base Model:} LLaMa3 8B
    \item \textbf{Individual defences:}
    \begin{itemize}
        \item LLaMa3 8B + entropy-based model
        \item LLaMa3 8B + uncertainty-based model
        \item LLaMa3 8B + geometric-based model
       
    \end{itemize}
    \item \textbf{Hybrid Adversarial Defence Models}
    \begin{itemize}
        \item LLaMa3 8B + LWV Fusion 
        \item LLaMa3 8B + ML Routing 
    \end{itemize}
\end{itemize}

We use four benchmark NLU datasets covering the range of reasoning types shown in Table~\ref{tab:exp1_datasets} and described in Section~\ref{dataset_spec}. All models are fine-tuned on in-domain training data using LoRA (rank 8, 0.1\% trainable parameters) and evaluated using corresponding test sets. Model code and hyperparameters for all our experiments are released as open source \footnote{GITHUB URI redacted during review period} for reproducibility purposes. 

\begin{table}[H]
\centering
\small
\begin{tabular}{|l|l|r|r|}
\hline
\textbf{Dataset} & \textbf{Domain} & \textbf{Samples}\\

\hline
FEVER & Fact Verification & 33,809 \\
HotpotQA & Multi-hop & 7 405  \\
CSQA & Commonsense & 9,741  \\
SIQA & Social Intelligence & 33,410  \\

\hline
\end{tabular}
\caption{In-domain training and evaluation datasets}
\label{tab:exp1_datasets}
\end{table}

Per-domain entropy thresholds used by Entropy-Based Model were tuned on validation splits with final values: FEVER (1.62), HotpotQA (1.84), CSQA (2.04), and SIQA (1.94). We employ TextAttack recipes shown in Table~\ref{tab:exp1_attacks} with fixed random seed for reproducibility and consistency across all model evaluations. All experiments were conducted using a single NVIDIA A100 GPU (80GB memory). Fine-tuning used LoRA (rank 8) for parameter efficiency (0.1\% trainable parameters), Adam optimizer, and a learning rate of $2 \times 10^{-5}$.

\begin{table}[H]
\small
\begin{tabular}{|l|p{4.5cm}|}
\hline
\textbf{Attack } & \textbf{Mechanism} \\
\hline
Character-level & Controlled typographic perturbations preserving semantic meaning \\
TextFooler & Embedding-based synonym substitution using cosine similarity \\
TextBugger & Homoglyph replacement and character-level manipulation \\
PWWS & WordNet-based semantic substitution with grammatical preservation \\
\hline
\end{tabular}
\caption{\citet{morris2020textattack} TextAttack framework adversarial attack strategies used for robustness evaluation}
\label{tab:exp1_attacks}
\end{table}

\subsection{Evaluation Metrics}

For clean data, without adversarial perturbations, we report NLU task accuracy and F1 scores to evaluate standard task performance. Since some models can refuse to answer, we report a refusal rate (Ref.) alongside the accuracy (Acc.) and F1 metrics, which are computed only on test data that was not refused.

For adversarially attacked data, we report NLU task accuracy and F1 score on non-refused test data, along with defence-specific metrics including refusal rate and attack success rate (ASR). Refusal rate measures the proportion of test data for which the model refused to answer, while attack success rate measures the proportion of adversarial inputs that result in incorrect predictions.

\subsection{In-Domain Task Performance}

%```latex
\begin{table*}[h]
\centering
\scriptsize
\setlength{\tabcolsep}{2pt}
\renewcommand{\arraystretch}{0.95}

\small
\begin{tabular}{|l|ccc|ccc|ccc|ccc|}
\hline
& \multicolumn{3}{c|}{\textbf{FEVER}} &
\multicolumn{3}{c|}{\textbf{HotpotQA}} &
\multicolumn{3}{c|}{\textbf{CSQA}} &
\multicolumn{3}{c|}{\textbf{SIQA}} \\
\textbf{Model} 
& \textbf{Acc.$\uparrow$} & \textbf{F1$\uparrow$} & \textbf{Ref.$\downarrow$}
& \textbf{Acc.$\uparrow$} & \textbf{F1$\uparrow$} & \textbf{Ref.$\downarrow$}
& \textbf{Acc.$\uparrow$} & \textbf{F1$\uparrow$} & \textbf{Ref.$\downarrow$}
& \textbf{Acc.$\uparrow$} & \textbf{F1$\uparrow$} & \textbf{Ref.$\downarrow$} \\
\hline

\multicolumn{13}{|c|}{\textit{Baselines}} \\
\hline

Base Model (no defence) &
51.76 & 50.23 & N/A &
44.00 & 45.20 & N/A &
68.14 & 68.72 & N/A &
52.60 & 51.30 & N/A \\

Entropy-Based Model &
59.75 & 58.92 & TBC &
46.60 & 47.80 & TBC &
78.50 & 76.20 & TBC &
64.30 & 62.59 & TBC \\

Uncertainty-Based Model &
64.48 & 63.71 & TBC &
69.10 & 68.50 & TBC &
75.10 & 74.80 & TBC &
70.10 & 70.90 & TBC \\

Geometric-Based Model &
\underline{81.65} & \underline{82.65} & N/A &
79.26 & 79.45 & N/A &
78.62 & 78.71 & N/A &
79.99 & 80.12 & N/A \\

FreeLB &
76.33 & 77.82 & N/A &
46.77 & 47.32 & N/A &
74.11 & 70.41 & N/A &
82.14 & 79.64 & N/A \\
\hline

\multicolumn{13}{|c|}{\textit{Hybrid Adversarial Defence Routing Models}} \\
\hline

LWV &
81.13 & 79.15 & 28.72 &
\underline{83.51} & \textbf{82.14} & \textbf{11.78} &
\underline{88.78} & \underline{87.45} & 11.22 &
\underline{90.63} & \underline{89.23} & 9.37 \\

ML &
\textbf{95.10} & \textbf{89.56} & \textbf{14.11} &
\textbf{88.02} & \underline{80.68} & 22.15 &
\textbf{89.35} & \textbf{87.92} & \textbf{0.08} &
\textbf{91.61} & \textbf{90.18} & \textbf{0.31} \\
\hline

\end{tabular}

\caption{In-domain clean task performance across all datasets without adversarial attacks. Accuracy and F1 are computed only for non-refused test data, with refusal rates reported for each model. A refusal rate of N/A means model answers all test data. 
\textbf{Bold} = best result, \underline{underline} = second best.}
\label{tab:clean_merged}
\end{table*}
%```

Results in table~\ref{tab:clean_merged} show task performance on clean test data without adversarial perturbations across all four datasets. Baseline defence methods consistently improve over the undefended model, with the geometric-based approach providing the strongest individual performance. Overall, ML routing achieves the highest accuracy across all datasets, where it improves performance by approximately 8.8 to 13.5 percentage points depending on the dataset. By allowing models to refuse to answer test data they are not sure about, our hybrid methods achieve a substantial reduction in hallucinations at the expense of some test data recall. To put the in-domain clean task results into a wider context, the latest non-adversarial agentic LLM methods such as LLM-Wiki \citep{ming2026retrievalreasoningselfevolvingagentnative} report an F1 score of 83.9 for HotpotQA across the entire test data.

\begin{table*}[h]

\centering
\scriptsize
\setlength{\tabcolsep}{2pt}
\renewcommand{\arraystretch}{0.95}

\small
\begin{tabular}{|l|cccc|cccc|cccc|cccc|}
\hline
&
\multicolumn{4}{c|}{\textbf{FEVER}} &
\multicolumn{4}{c|}{\textbf{HotpotQA}} &
\multicolumn{4}{c|}{\textbf{CSQA}} &
\multicolumn{4}{c|}{\textbf{SIQA}} \\
\textbf{Model} &
\rotatebox{90}{Acc.$\uparrow$} & \rotatebox{90}{F1$\uparrow$} & \rotatebox{90}{Ref.$\uparrow$} & \rotatebox{90}{ASR$\downarrow$} &
\rotatebox{90}{Acc.$\uparrow$} & \rotatebox{90}{F1$\uparrow$} & \rotatebox{90}{Ref.$\uparrow$} & \rotatebox{90}{ASR$\downarrow$} &
\rotatebox{90}{Acc.$\uparrow$} & \rotatebox{90}{F1$\uparrow$} & \rotatebox{90}{Ref.$\uparrow$} & \rotatebox{90}{ASR$\downarrow$} &
\rotatebox{90}{Acc.$\uparrow$} & \rotatebox{90}{F1$\uparrow$} & \rotatebox{90}{Ref.$\uparrow$} & \rotatebox{90}{ASR$\downarrow$} \\
\hline

\multicolumn{17}{|c|}{\textit{Baselines}} \\
\hline
Base Model (no defence) &
24.46 & 36.20 & N/A & 75.50 &
50.99 & 49.26 & N/A & 52.62 &
48.03 & 48.03 & N/A & 63.12 &
40.60 & 50.41 & N/A & 63.45 \\

Entropy-Based Model &
58.10 & 58.10 & 56.38 & 43.62 &
59.90 & 58.92 & 47.38 & 41.01 &
62.90 & 55.12 & 36.88 & 51.97 &
61.50 & 56.77 & 76.81 & 23.19 \\

Uncertainty-Based Model &
63.45 & 63.45 & 43.00 & 58.00 &
56.90 & 41.47 & 22.48 & 47.52 &
68.60 & 68.70 & 34.33 & 45.67 &
62.55 & 62.55 & 62.59 & 37.45 \\

Geometric-Based Model &
76.25 & \textbf{77.25} & N/A & 33.20 &
76.30 & 74.53 & N/A & 25.59 &
70.54 & 67.28 & N/A & 27.53 &
75.51 & 72.81 & N/A & 22.87 \\

FreeLB &
67.29 & 67.29 & N/A & 42.71 &
43.76 & 43.76 & N/A & 60.95 &
22.11 & 22.11 & N/A & 77.89 &
65.03 & 65.03 & N/A & 34.97 \\
\hline

\multicolumn{17}{|c|}{\textit{Hybrid Adversarial Defence Routing Models}} \\
\hline
LWV &
\underline{77.35} & \underline{76.43} & \underline{70.09} & \underline{29.91} &
\underline{75.21} & \underline{76.48} & \underline{76.42} & \underline{23.48} &
\underline{87.02} & \underline{86.77} & \underline{83.61} & \underline{17.47} &
\underline{93.64} & \underline{93.76} & \textbf{97.92} & \textbf{2.75} \\

ML &
\textbf{89.38} & 76.25 & \textbf{81.11} & \textbf{13.23} &
\textbf{79.85} & \textbf{81.20} & \textbf{89.96} & \textbf{10.00} &
\textbf{87.20} & \textbf{86.96} & \textbf{86.47} & \textbf{14.47} &
\textbf{94.06} & \textbf{94.15} & \underline{96.81} & \underline{3.86} \\
\hline
\end{tabular}

\caption{In-domain robustness under adversarial attack using TextAttack across all datasets. Accuracy and F1 are computed only for non-refused test data, with refusal rates reported for each model. A refusal rate of N/A means model answers all test data.
\textbf{Bold} = best result, \underline{underline} = second best.}
\label{tab:adversarial_final}
\end{table*}

Results in table \ref{tab:adversarial_final} shows performance under adversarial attack, with the geometric-based model achieving the strongest
individual robustness among baseline methods. ML Routing outperforms LWV, and achieves a 3.5\% to 19\% accuracy improvement over the best baseline model and a 13\% to 20\% reduction in ASR.

A detailed breakdown of base model vulnerability by attack type is provided in Appendix~\ref{appendix:attack_breakdown}.

\subsection{Discussion}

Traditional adversarial training (FreeLB) improves robustness through
exposure to perturbed examples during training. However, it exhibits
three key limitations: (1) it cannot refuse inputs, leaving no fallback
for novel attack patterns; (2) it incurs substantial training cost while
remaining weaker than hybrid approaches; and (3) its transfer from
continuous embedding perturbations to discrete token-level attacks is
inconsistent, as illustrated by its sharp performance drop on CSQA.

The uncertainty-based model depends on internal confidence estimations, which lead to conservative behavior (high rejection rate).  The entropy-based model is more selective, reducing false positives but potentially missing a larger proportion of attacks. The geometric-based model works in embedding space and detects the attacks without making significant changes in confidence and entropy. This complementarity explains the effectiveness of the hybrid methods.

An additional observation is the improvement in clean task performance. The geometric-based model appears to contribute useful structural features even without adversarial perturbations, while the uncertainty-based model improves decision calibration by learning when to abstain. This suggests that the proposed defence mechanisms actively reduce hallucination.

Both approaches combine the same three experts but differ in how predictions are aggregated. LWV performs soft averaging, while ML Routing selects a single expert per instance. The advantage of ML Routing is most pronounced when geometric features are strong, where hard selection avoids diluting informative features. When features are less separable, the difference between the two approaches is smaller, suggesting that soft averaging remains competitive in such cases.

\section{OOD Adversarial Defence}
\label{sec:exp2}

\begin{table}[H]
\centering
\small
\begin{tabular}{|l|l|r|}
\hline
\textbf{Dataset} & \textbf{Domain} & \textbf{Samples} \\
\hline
AERO & Aerospace Engineering & 80 \\
CPIQA & Climate Science & 2,364 \\
\hline
\end{tabular}
\caption{OOD evaluation datasets}
\label{tab:exp2_datasets}
\end{table}

\begin{table*}[t]
\centering
\setlength{\tabcolsep}{3pt}
\renewcommand{\arraystretch}{0.95}
\small

\begin{tabular}{|l|ccc|ccc|cccc|cccc|}
\hline
& \multicolumn{6}{c|}{\textbf{Clean}} & \multicolumn{8}{c|}{\textbf{Adversarial Attack}} \\
& \multicolumn{3}{c|}{\textbf{AERO}} 
& \multicolumn{3}{c|}{\textbf{CPIQA}} 
& \multicolumn{4}{c|}{\textbf{AERO}} 
& \multicolumn{4}{c|}{\textbf{CPIQA}} \\

\textbf{Model} 
& \textbf{Acc.$\uparrow$} & \textbf{F1$\uparrow$} & \textbf{Ref.$\downarrow$}
& \textbf{Acc.$\uparrow$} & \textbf{F1$\uparrow$} & \textbf{Ref.$\downarrow$}
& \textbf{Acc.$\uparrow$} & \textbf{F1$\uparrow$} & \textbf{Ref.$\uparrow$} & \textbf{ASR$\downarrow$}
& \textbf{Acc.$\uparrow$} & \textbf{F1$\uparrow$} & \textbf{Ref.$\uparrow$} & \textbf{ASR$\downarrow$} \\
\hline

\multicolumn{15}{|c|}{\textit{Baselines}} \\
\hline

Base Model 
& 49.00 & 47.34 & N/A 
& 45.00 & 43.86 & N/A
& 45.00 & 46.33 & N/A & 65.70
& 27.00 & 27.00 & N/A & 78.80 \\

FreeLB 
& 62.52 & 61.84 & N/A
& 48.00 & 55.97 & N/A
& 45.45 & 54.55 & N/A & 50.55
& 34.20 & 40.27 & N/A & 70.94 \\

\hline
\multicolumn{15}{|c|}{\textit{Hybrid Adversarial Defence Routing Models}} \\
\hline

LWV
& \textbf{92.50} & \textbf{91.53} & TBC
& \underline{82.11} & \underline{79.10} & TBC
& \underline{78.33} & \underline{78.33} & \underline{54.37} & \underline{45.20}
& \underline{80.43} & \underline{80.43} & \underline{58.80} & \underline{42.12} \\

ML
& \textbf{92.50} & \underline{89.50} & TBC
& \textbf{91.37} & \textbf{90.70} & TBC
& \textbf{85.00} & \textbf{85.00} & \textbf{62.00} & \textbf{33.00}
& \textbf{84.14} & \textbf{84.14} & \textbf{79.11} & \textbf{25.89} \\

\hline
\end{tabular}

\caption{Out-of-distribution clean and adversarial performance. Accuracy and F1 are computed only for non-refused test data, with refusal rates reported for each model. A refusal rate of N/A means model answers all test data. \textbf{Bold} = best, \underline{underline} = second best.}
\label{tab:ood_merged}
\end{table*}
To assess how adversarially defended models generalize beyond their training data distribution, we extend the analysis to out-of-distribution (OOD) datasets shown in Table~\ref{tab:exp2_datasets}. All models are fine-tuned on an aggregation of all in-domain training data and evaluated on the OOD testsets. All OOD adversarial evaluations use the same aggregated attack strategy as the in-domain setup. Results in Table~\ref{tab:ood_merged} show model performance for both clean and adversarially attacked OOD data. 

\subsection{Discussion}

The base model behaves quite differently across the two OOD domains. While AERO maintains 45\% accuracy under attack, CPIQA drops sharply to 27\%. Our hypothesis for this is that CPIQA contains more flexible natural language, making synonym-based attacks easier, whereas AERO relies on more constrained technical vocabulary. FreeLB shows limited transfer abilities to OOD data. This is likely due to the mismatch between continuous perturbations used in training and discrete token-level attacks used at evaluation.

The best performance is achieved by ML Routing which reduced attack success to 33\% on AERO and 25.89\% on CPIQA. This improvement comes from both stronger predictions and better attack detection, reflected in higher refusal rates.

\section{Prompt Injection and Jailbreak Defence}
\label{sec:exp3}

Beyond traditional adversarial perturbations, modern LLM threats include \textit{prompt injection} and \textit{jailbreaking} attacks. These attacks exploit instruction-following behavior rather than lexical perturbations, representing a fundamentally different threat model. We evaluate the same hybrid defence models as our previous experiments to assess zero-shot generalization to prompt-level attacks alongside our baseline models.

\begin{table}[H]
\centering
\small
\setlength{\tabcolsep}{4pt}
\begin{tabular}{|l|p{3cm}|c|c|}
\hline
\textbf{Dataset} & \textbf{Attack Type} & \textbf{Samples} & \textbf{Threshold}\\
\hline
Safe-guard & Prompt injection & 10,296 & 4.60\\
AdvBench & Jailbreaking & 520 & 2.50\\
DAN & Jailbreaking & 1,405 & 0.6709\\
\hline
\end{tabular}
\caption{Prompt injection and jailbreaking evaluation datasets.}
\label{tab:exp3_datasets}
\end{table}

\begin{table*}[t]
\centering
\setlength{\tabcolsep}{4pt}
\renewcommand{\arraystretch}{0.95}
\small

\begin{tabular}{|l|cccc|cc|cc|}
\hline
& \multicolumn{4}{c|}{\textbf{SafeGuard}} & \multicolumn{2}{c|}{\textbf{AdvBench}} & \multicolumn{2}{c|}{\textbf{DAN}} \\
\textbf{Model} 
& \textbf{Acc.$\uparrow$} & \textbf{F1$\uparrow$} & \textbf{Ref.$\uparrow$} & \textbf{ASR$\downarrow$}
& \textbf{Ref.$\uparrow$} & \textbf{ASR$\downarrow$}
& \textbf{Ref.$\uparrow$} & \textbf{ASR$\downarrow$} \\
\hline

\multicolumn{9}{|c|}{\textit{Baselines}} \\
\hline

ProtectAI v1 
& \underline{81.53} & 57.24 & 40.79 & 59.21
& 0.58 & 99.42
& 73.67 & 26.33 \\

NeMoGuard    
& 45.22 & 47.62 & 21.27 & 78.73
& 85.58 & 14.42
& \textbf{96.44} & \textbf{3.56} \\

\hline
\multicolumn{9}{|c|}{\textit{Hybrid Adversarial Defence Routing Models}} \\
\hline

LWV        
& 58.16 & \underline{58.16} & \textbf{91.79} & \textbf{8.21}
& \textbf{97.50} & \textbf{2.50}
& 77.72 & 22.28 \\

ML 
& \textbf{82.65} & \textbf{82.65} & \underline{90.38} & \underline{9.62}
& \underline{97.31} & \underline{2.69}
& \underline{85.69} & \underline{14.31} \\

\hline
\end{tabular}

\caption{Zero-shot evaluation of hybrid models on prompt injection (SafeGuard) and jailbreak (AdvBench, DAN) datasets.
\textbf{Bold} = best, \underline{underline} = second best.}
\label{tab:cross_domain_evaluation}
\end{table*}

\begin{table*}[h]
  \centering
  \small
  \begin{tabular}{|l|p{6.5cm}|c|p{4.5cm}|}
    \hline
    \textbf{Attack Recipe} & \textbf{Mechanism} & \textbf{$\Delta H$ (mean)} & \textbf{Example Transform} \\
    \hline
    \texttt{character\_level} & Neighboring character swaps, insertions, deletions & +0.12 & ``playing'' $\rightarrow$ ``plaaying'' (neighbor swap/insert) \\
    \hline
    \texttt{textfooler\_simple} & Embedding-based word substitutions constrained by semantic similarity & +0.55 & ``movie'' $\rightarrow$ ``film'' (embedding word-swap) \\
    \hline
    \texttt{textbugger\_simple} & Mixed character and word-level perturbations (homoglyphs, char edits + embedding swaps) & +0.30 & ``the'' $\rightarrow$ ``th3'' / ``colour'' $\rightarrow$ ``color'' \\
    \hline
    \texttt{pwws\_simple} & WordNet-based synonym substitutions (semantic-presing swaps) & +0.48 & ``happy'' $\rightarrow$ ``glad'' (WordNet swap) \\
    \hline
  \end{tabular}
  \caption{TextAttack recipes used for adversarial generation, their primary perturbation mechanisms, observed mean token-level entropy change ($\Delta H$) and example transformations.}
  \label{tab:attack_strategies}
\end{table*}

We use one prompt injection and two jailbreakings benchmark datasets, shown in Table~\ref{tab:exp3_datasets} and described in Section~\ref{dataset_spec}.

\subsection{Discussion}

For prompt injection ML Routing has the best overall performance, with a balance between accuracy (82.65\%) and refusal (90.38\%). ProtectAI v1 produced high precision but low recall, missing a large percentage of attacks, which is consistent with \citet{hackett2025}, who measured a detection rate of 38.31\%, closely aligned with the 40.79\% refusal. NeMoGuard showed the opposite pattern, with high recall but low precision, indicating that it was not trained on prompt injection data and, therefore, had a difficult time distinguishing between benign and malicious input.

For jailbreaking, ML Routing provided the most consistently high overall performance, maintaining strong refusal rates without the instability observed in other methods. NeMoGuard had the highest refusal rate of 96.44\% for the DAN dataset, which we hypothesize that it was trained with data containing structurally similar jailbreaks to the DAN dataset. ProtectAI v1 performs very poorly (0.58\% refusal) on the AdvBench dataset which we think is due to task misalignment, since it is fine-tuned for prompt injection patterns which are absent in AdvBench. These results highlight the advantage of hybrid models which combine complementary features, enabling adversarial defence based on model behaviour rather than surface form.

\section*{Conclusion}

We present a hybrid adversarial defence framework that combines entropy-based detection, uncertainty-aware training, and geometric representation enhancement. Across multiple NLU datasets our results show that integrating these complementary features improves both clean task performance, reducing hallucinations, and performance under adversarial attack, increasing robustness.

Our hybrid model uses features that are invariant across domains, providing an approach that can generalize to any domain.

For future work we would like to explore adaptive attack scenarios which target the whole hybrid pipeline. Attackers could theoretically exploit vulnerabilities arising from both expert model and the selector network.

Another future direction would be to investigate a unified multi-head defense architecture. In contrast to training different experts and routing among them, a single Mixture of Experts type model could learn specialized attention heads focusing on the feature types identified in this paper. This would allow removal of the routing model completely and provide a simpler end to end model.

\section*{Limitations}

All experiments use a base LLM of LLaMa3-8B. We see no theoretical reason why our results will not generalize to other base LLMs but we leave it as future work to confirm this.

The LWV and ML Routing models work as a trainable black-box routers. This might create problems for safety-critical deployments that demand auditable and/or explainable decision processes.

\appendix

\section{Feature Vector Description}
\label{appendix:features}

The hybrid aggregation model features are organised into six categories:

\textbf{Textual features (6):} word count, character count, lexical diversity (unique / total words), punctuation density, whitespace and uppercase ratio.

\textbf{Entropy-based features (6):} entropy $H = -\sum p \log p$, threshold distance $|H - \theta_d|$, confidence derived from entropy, a flag indicating whether $H > \theta_d$, token-level perplexity, and semantic shift between original and perturbed inputs.

\textbf{Uncertainty-based features (7):} answer confidence, certainty score (``I am sure'' probability), combined confidence, uncertainty ($1 - \text{confidence}$), a certainty flag (sure/unsure), margin from the decision boundary, and confidence gap.

\textbf{Domain encoding (5):} binary indicators for FEVER, CSQA, SIQA, HotpotQA, and OTHER.

\textbf{Cross-expert features (4):} confidence difference between experts, agreement flag, minimum confidence, and maximum confidence.

\textbf{Geometric-based features (3):} purification magnitude $\|\mathbf{h} - \mathbf{h}'\|_2$, norm ratio $\|\mathbf{h}'\|_2 / \|\mathbf{h}\|_2$, and SVD dominance $\sigma_1 / \sum_i \sigma_i$.

%\FloatBarrier % prevent defence model fig appearing before the experiment section

\section{Entropy-based Model TextAttack Recipes}
\label{appendix:attack_strategies}
Table~\ref{tab:attack_strategies} shows the TextAttack recipes used.

\section{Base Model Vulnerability Analysis by Attack Type}
\label{appendix:attack_breakdown}

\begin{table}
\centering
\small
\setlength{\tabcolsep}{4pt}
\begin{tabular}{|l|c|c|c|c|}
\hline
\textbf{Attack Type} & \textbf{Acc} & \textbf{F1-Mi} & \textbf{F1-Ma} & \textbf{Att. Succ.} \\
\hline
Character-level & 11.72 & 11.72 & 10.78 & 88.28 \\
TextFooler      & 35.04 & 35.04 & 34.37 & 64.96 \\
TextBugger      &  2.36 &  2.36 &  2.31 & 97.64 \\
PWWS            & 20.37 & 20.37 & 20.24 & 79.63 \\
\hline
\end{tabular}
\caption{Base model vulnerability by attack type on FEVER dataset.}
\label{tab:exp1_fever_base_attacks}
\end{table}

Table~\ref{tab:exp1_fever_base_attacks} presents results for the
undefended LLaMA-3-8B base model when exposed to four adversarial attack families on FEVER dataset. The base model refuses none of the adversarial inputs (0\% refusal), confirming it operates without defence mechanisms. These results establish the reference point for understanding main results in (Table~\ref{tab:adversarial_final}).

%
%\begin{table}
%\centering
%\small
%\setlength{\tabcolsep}{4pt}
%\begin{tabular}{|l|c|c|c|}
%\hline
%\textbf{Dataset} & \textbf{Entropy} & %\textbf{Uncertainty} & \textbf{Geometric} \\
%\hline
%SafeGuard & 48.56 & 45.00 & 6.44 \\
%\hline
%AdvBench  & 10.00 & 68.26 & 21.74 \\
%\hline
%DAN       & 0.00  & 45.10 & 54.90 \\
%\hline
%\end{tabular}
%\caption{Routing distribution across expert models (\%).}
%\label{tab:routing_distribution}
%\end{table}

\clearpage

\section*{Acknowledgments}
This work was supported by the Economic and Social Research Council (ES/V011278/1). The authors acknowledge the IRIDIS High-Performance Computing Facility at the University of Southampton.
\nocite{*}
\bibliography{custom}

\end{document}